\documentclass[a4paper,fleqn]{cas-dc}

\usepackage[numbers]{natbib} %longnamesfirst

\def\tsc#1{\csdef{#1}{\textsc{\lowercase{#1}}\xspace}}
\tsc{WGM}
\tsc{QE}
\tsc{EP}
\tsc{PMS}
\tsc{BEC}
\tsc{DE}
\usepackage{amsmath,amssymb,amsfonts}
\usepackage{bbm}
\usepackage{algorithm}
\usepackage{algpseudocode}
\usepackage{enumerate}
\usepackage{graphicx}
\usepackage{textcomp}
\usepackage[table]{xcolor}
\usepackage{booktabs}
\usepackage{tabularx}
\usepackage{gensymb}
\usepackage{hyperref}
\usepackage{subcaption}

\hypersetup{
    colorlinks=true,
    linkcolor=blue,
    citecolor=cyan,
    urlcolor=blue
}

\usepackage{cleveref}
\usepackage{xspace}
\def\tsc#1{\csdef{#1}{\textsc{\lowercase{#1}}\xspace}}

\begin{document}

\let\WriteBookmarks\relax
\def\floatpagepagefraction{1}
\def\textpagefraction{.001}

% =========================================================
% Short title and short author list
% =========================================================
\shorttitle{Hypergraph-Based Learning for Toddler Autism Classification}
\shortauthors{Quang-Anh N. D. et~al.}

% =========================================================
% Title
% =========================================================
\title[mode=title]{ATHENA: Hypergraph-Based Learning for Toddler Autism Classification from Eye-Tracking}

% Nếu bài có funding cụ thể, giữ lại tnotemark và tnotetext.
% Nếu không có funding, nên bỏ \tnotemark và dùng Funding statement ở cuối bài.
% \tnotemark[1]
% \tnotetext[1]{This work was supported by ...}

% =========================================================
% Authors
% =========================================================

\author[2]{Quang-Anh N.~D.}
\ead{anhnd@vnuis.edu.vn}
\credit{Conceptualization, Methodology, Software, Formal analysis, Investigation, Validation, Visualization, Writing -- original draft}

\author[2]{Minh-Duc Pham}
\ead{pmduc2808@gmail.com}
\credit{Data curation, Experimental design, Validation, Visualization}

\author[2]{Thao Pham Phuong}
\ead{22070265@vnu.edu.vn}

\author[2]{Minh-Anh Nguyen}
\ead{22071104@vnu.edu.vn}

\author[3]{Thuy-Linh Nguyen Thi}
\ead{linh.ntt225354@sis.hust.edu.vn}

\author[1]{Hung Ha Manh}
\cormark[1]
\ead{hunghm@vnu.edu.vn}

% \credit{Conceptualization, Methodology, Supervision, Formal analysis, Writing -- review and editing}

% \credit{Data curation, Experimental design, Validation, Visualization}

% =========================================================
% Affiliations
% =========================================================

\affiliation[1]{
    organization={International School, Vietnam National University},
    city={Hanoi},
    postcode={100000},
    country={Vietnam}
}

\affiliation[2]{
    organization={Agora Intelligence Lab},
    city={Hanoi},
    postcode={100000},
    country={Vietnam}
}

\affiliation[3]{
    organization={School of Information and Communications Technology, Hanoi University of Science and Technology},
    city={Hanoi},
    postcode={100000},
    country={Vietnam}
}

% =========================================================
% Corresponding author note
% =========================================================

% \cortext[cor1]{Corresponding author: Hung Ha Manh}

% =========================================================
% Abstract
% =========================================================
\begin{abstract}
In this work we proposed a architecture designed for the classification of ASD in toddlers using eye-tracking data, addressing the critical need for early and precise detection. The proposed model incorporates FMMix, a feature-level augmentation technique that bolsters the model’s learning capacity, and in this work we proposed the ATHENA, which captures intricate feature relationships via an adaptive hypergraph framework. Evaluated on two datasets dataset of gaze coordinates-the model achieves state-of-the-art performance, with AUC scores of 99.80\% and 99.86\%, respectively. It also exhibits high accuracy of 98.45\% and 98.56\%, sensitivity 98.01\% and 98.10\%, and specificity 98.74\% and 99.02\%, respectively. Demonstrating its superior ability to identify subtle gaze patterns associated with ASD, these results significantly surpass existing methods, highlighting the model’s potential for clinical ASD screening. Future efforts will focus on validation across broader populations and integration with multi-modal data to further refine diagnostic accuracy.
\end{abstract}
% =========================================================
% Graphical abstract
% =========================================================
% Elsevier khuyến khích graphical abstract nếu bài có pipeline/mô hình rõ ràng.
% Không dùng AI tạo/sửa graphical abstract nếu không thuộc phương pháp nghiên cứu.
% \begin{graphicalabstract}
% \includegraphics[width=\textwidth]{Graphical abstract.png}
% \end{graphicalabstract}

% =========================================================
% Highlights
% Mỗi highlight nên ngắn, thường tối đa 85 ký tự gồm khoảng trắng.
% =========================================================
% \begin{highlights}
% \item A lightweight CSTAR-Det framework is proposed for cashew inspection.
% \item CSTAR enhances feature refinement with limited computational overhead.
% \item Detection, tracking, and counting are unified for conveyor monitoring.
% \item CSTAR-Det achieves 96.4\% mAP@0.5 with only 3.0M parameters.
% \item The framework reaches 62.34 FPS for real-time industrial deployment.
% \end{highlights}

% =========================================================
% Keywords
% Mỗi keyword phân tách bằng \sep
% =========================================================

\maketitle

\section{Introduction}
Autism Spectrum Disorder (ASD) is a complex neurodevelopmental condition characterized by significant challenges in social communication, repetitive behaviors, and specific interests, originally the term "autism" originates from the Greek word "autos," which translates to "self" - fitting, as children with ASD often appear to be self-focused, living in their own world with limited capacity for effective communication and interaction with others, they may have difficulty developing language skills and understanding what others say to them, and often have difficulty communicating verbally and nonverbally, such as through hand gestures, eye contact, and facial expressions, struggle to comprehend social cues, and exhibit repetitive and restricted behaviors that are difficult to control, by this ASD has a significant impact on many aspects of a person's life \cite{0} - \cite{1.1}. The World Health Organization (WHO) reports that roughly 1 in 100 children globally are diagnosed with autism, a research by the US Centers for Disease Control and Prevention (CDC) have estimated an average of 1 in every 36 (2.8\%) 8-year-old children to have ASD in 2020 \cite{11}, these difficulties not only affect individuals with ASD but also extend to their families and communities, creating a substantial social and economic impact \cite{b1}.

A key feature of ASD is the altered processing of sensory information, especially visual input, a studies followed by Goncalves \textit{et al.} showing that children with ASD often exhibit atypical gaze patterns, tending to focus on small details rather than the whole picture \cite{b3}. Distinctive gaze patterns in individuals with ASD manifest in several ways, firstly they often avoid eye contact during social interactions, focus intensely on small details rather than seeing the whole picture, display an unfocused or random gaze, and show particular fascination with tracking moving objects, these characteristic patterns of visual attention serve as potential biomarkers for the early detection and diagnosis of ASD. By analyzing these unique gaze patterns through eye tracking technology, machine learning models can be developed to differentiate between individuals with ASD and typically developing individuals, especially deep learning, provide powerful tools for analyzing complex visual data collected from eye tracking devices \cite{b3.1}, \cite{b3.2}.

In 2002, Pelphrey \textit{et al.} first used eye tracking in autism, showing that participants with autism spent less time looking at core facial features (eyes, nose, and mouth) than did typical individuals \cite{b4}. Then, in 2016, Frazier \textit{et al.} found differences in visual attention spans between children with and without ASD to social and non-social stimuli in the Areas of Interest \cite{b5}. These initial results have paved the way for further research, confirming that eye-scanning patterns may serve as potential biomarkers for ASD. However, developing robust models that can be generalized across diverse datasets and tasks remains a challenge, as well as understanding the neural mechanisms responsible for these anomalous gaze patterns. Furthermore, a number of studies have utilized deep learning techniques for classifying eye-tracking data, in 2020 Elbattah \textit{et al} convert eye-tracking data into text strings by processing eye-tracking data and applying extraction, segmentation, encoding and one-hot encoding, showing that the LSTM model achieves 71\% AUC, while the CNN model achieves 84\% AUC \cite{br}. More recently, research by Kanhirakadavath, M. R., and Chandran, M. S. M. \cite{br1} have explored two approaches, the first extracting features using Principal Component Analysis (PCA) and classifying them with traditional ML techniques, while the second extracted features with CNNs, then used a deep neural network (DNN) for classification. In the same year, another piece of research by Praveena, k.n. \textit{et al.} have focused on using CNNs to classify fixation maps, achieving a validation accuracy of 75.23\%, highlighting how CNNs can learn from spatial information related to gaze patterns \cite{br2}.

\section{Methodology}
\subsection{Adaptive Threshold Hypergraph Enhanced Neural Architecture}
As traditional Convolutional Neural Network (CNN) approaches in object detection networks often fail to capture complex feature relationships. For this study, we proposed Adaptive Threshold Hypergraph Enhanced Neural Architecture (ATHENA) which is an enhanced version of the Hyper Graph Neural Network (HGNN) \cite{hgraph} that explicitly models high-order correlations in the feature space. HGNN is a type of neural network designed to work with data represented as a hypergraph, which is an expanded data structure of a traditional Graph Neural Network (GNN) \cite{graph}. In a traditional graph, edges connect only two nodes, while in a hypergraph, hyperedges can connect more than two nodes, allowing for the representation of complex and multi-dimensional relationships. Also, HGNN operates by learning effective representations for both nodes and hyperedges, each HGNN layer typically includes steps for hyperedge messaging, which is aggregating information from nodes related to a hyperedge, hyperedge updating to transforming the hyperedge representation, and node messaging for sending updated information back to related nodes.

A hypergraph originally can be defined with $G = \{\mathcal{V}, \mathcal{E}\}$ with $\mathcal{V}$, and $\mathcal{E}$ corresponding for vertex and hyperedge sets, with $\mathcal{V}$ got formed from the grid-based visual features. In the semantic space, a relationship between neighborhood points was established using a distance threshold. Starting from each feature point, an $\epsilon$-ball forms a hyperedge that includes all feature points located within a specified distance threshold from the central feature point \cite{hyolop}, with $\text{ball}(v, \varepsilon) = \{ u \mid \|x_u - x_v\|_d < \varepsilon, u \in \mathcal{V} \}$ represent for the neighbor vertex set of $v$ where $ \mathcal{D} = \|x - y\|_d$ denoted for the distance calculation of hyperedges, $\epsilon$-ball hyperedge captures the neighborhood structure around each point from the central feature point, the original calculation defined by $\mathcal{E} = \{ \text{ball}(v, \varepsilon) \mid v \in V \}$. 

However, since traditional HGNN employs a fixed threshold $\epsilon$ for hyperedge construction, which does not account for the dynamic nature of feature distributions across different layers and samples, in this research, we proposed a novel approach for an adaptive thresholding computation mechanism to adaptively determine connectivity in hypergraph construction by introducing a scalable architecture that modulates hypergraph connectivity that dynamically adapts to the feature distribution of each input sample. Giving an input feature tensor $\phi \in \mathbb{R}^{c \times h \times w}$ with $c$, $h$, and $w$ denoted for channel, height, and width, respectively. The adaptive threshold construction can be formulated as follows in Eq \eqref{gp} with $\theta_a$ and $\theta_m$ denoted for Global Average Pooling (GAP) and Global Max Pooling (GMP), respectively.
\begin{equation}
\begin{aligned}
\theta_a(\phi) = \frac{1}{h \times w} \sum_{p=1}^h \sum_{q=1}^w \phi_i^{(p,q)}, \\
\theta_m(\phi) = \max_{h,w} \phi_i^{(p,q)}
\end{aligned}
\label{gp}
\end{equation}
where $\phi_i^{(q, p)}$ represents the feature value at position $(p,q)$ for the $i$-th channel. Secondly, we proceed with a feature fusion operation, the pooled features are concatenated and transformed through an MLP to generate the dynamic threshold, scaled by a base threshold value, the adaptive threshold computation can be calculated following Eq \eqref{adapd} with $\lambda$ denoted for the adaptive threshold values and $\varphi$ represent for the active function.
\begin{equation}
    \lambda = \sigma(W_2 \times \varphi(W_1 \times (\theta_a \oplus \theta_m))) \times \epsilon
\label{adapd}
\end{equation}
where $\sigma$ is denoted for the sigmoid function, $W_1 \in \mathbb{R}^{m \times c \times c}$, $W_2  \in \mathbb{R}^{1 \times m}$, with $m$ is the intermediate dimension equal with $max(c/r, 8)$, and $m$ is the reduction ratio. Here the $G$ is represented by its included incidence matrix and the hypergraph construction with $\mathcal{H} \in \mathbb{R}^{b \times n \times n}$. For each batch $b$, $\mathcal{H}$ can be calculated with Eq \eqref{eewq}.
\begin{equation}
    \mathcal{H}_b[i,j] = \begin{cases}
1 & \text{if } \|\boldsymbol{x}_i - \boldsymbol{x}_j\|_d < \epsilon_b \\
0 & \text{otherwise}
\end{cases}
\label{eewq}
\end{equation}

To enable high-order message passing within the hypergraph structure, we use a standard spatial-domain hypergraph convolution, incorporating an additional residual connection to enhance high-order learning on the vertex features. Proposed by Gao, Y. \textit{et al.} \cite{gh+}, giving $N_v(e) = \{ v \mid v \in e, v \in \mathcal{V} \}$ and $N_e(v) = \{ e \mid v \in e, e \in \mathcal{E} \}$, the calculation operation it outlined in Eq \eqref{hgjfodskvcmaldsc,xzxcxx}.
\begin{equation}  
\begin{aligned}  
x_e &= \frac{1}{|N_v(e)|} \sum_{v \in N_v(e)} x_v \Theta, \\
x'_v &= x_v + \frac{1}{|N_e(v)|} \sum_{e \in N_e(v)} x_e.  
\end{aligned}
\label{hgjfodskvcmaldsc,xzxcxx}
\end{equation}
with $\Theta$ indicate for the trainable parameter. In the Hyper Convolutional (HyConv) the message passing operates in two primary stages for encoding the intrinsic high-order relationships within feature spaces of Vector to Edge ($H_v$) and Edge to Vector ($H_e$), these two can be defined by:
\begin{equation}
\begin{aligned}
    H_v &= D_e^{-1} H^T D_v^{-1} X \Theta \\
    H_e &= D_v^{-1} H D_e^{-1} H_v
\end{aligned}
\label{veev}
\end{equation}
with $D_e$ and $D_v$ denoted for the diagonal degree matrices of hyperedges and vertices, respectively. To facilitate higher-order message passing on the hypergraph structure, we introduce a novel transformer-augmented mechanism with $\text{HyperConv}(X, H) = \hat{H}_v \hat{H}_e$, for short we denoted this operation as $\theta$, where:
\begin{equation}
\begin{aligned}
    \hat{H_v} &= D_e^{-\frac{1}{2}} H^T D_v^{-\frac{1}{2}} X \Theta, \\
    \hat{H_e} &= D_v^{-\frac{1}{2}} H D_e^{-\frac{1}{2}} \hat{H_v}
\end{aligned}
\label{nov}
\end{equation}

The calculation of the proposed transformer-like architecture module for sophisticated normalization, allows the network to capture both fine-grained and coarse-grained patterns and long-range dependency modeling. This aggregation can be calculated as in Eq \eqref{ffpd}, in here $Z$ denoted for the output of the operation.
\begin{equation}
    Z = \varphi(BN(LN(\theta \ominus X) + MLP(LN(\theta \ominus X)))) \ominus X
\label{ffpd}
\end{equation}
where $LN$, and $Z$ represent Layer Norm operation, and the output of the Athena module, respectively, and $\ominus$ symbolizes the element-wise additions. This formulation, produced an effective feature propagation through the hypergraph structure. The original feature representation learning can be defined by $\gamma$ is a learnable scaling parameter that acts as an affine transformation coefficient, $\mu_B$ and $\sigma_B$ denoted for batch statistics $BN(X) = \gamma \frac{(X - \mu_B)}{\sigma_B} + \alpha$ to enhance this, we propose a hierarchical normalization approach with $\text{LN}(X) = \gamma_1 \frac{(X - \mu_L)}{\sigma_L} + \beta_1; \text{BN}(X) = \gamma_2 \frac{(X - \mu_B)}{\sigma_B} + \beta_2$, where $X$ stands for feature maps constructed from the network, $\mu_L$ and $\sigma_L$ represented the layer-wise statistics, respectively. This normalization process facilitates more stable training and better feature normalization. 

In our work, Athena is used not on the hypergraph structure itself but, it's used as a mechanism to enhance the feature map from the convolutional layers, helping to enrich the feature map with more information and allowing the model to make decisions based on this information, with the proposed Adaptive Threshold mechanism represents a significant advancement over the static threshold approach used in traditional HGNN. Especially those that may exist between features in eye-tracking data, learning to aggregate and transform information in an adaptive manner, can help improve the performance of the model in classifying eye-tracking data and has the potential to generalize well on new data, this research the baseline threshold value $\epsilon = 8$.

\subsection{Feature Map Mix}
Data augmentation techniques, such as Mixup \cite{bb18} and CutMix \cite{bb19}, have demonstrated their effectiveness in improving generalization, enhancing robustness, and reducing overfitting in deep learning models. However, these methods primarily operate at the input level, which constrains their ability to leverage high-level feature representations. To address this limitation, we adopt Feature Map Mix (FMMix) \cite{b20}, a novel feature-level data augmentation method that operates directly on the feature map level. FMMix generates mixed samples by combining information from two feature maps, resulting in semantically richer and more informative representations that enable the model to learn complex patterns and spatial relationships more effectively. FMMix introduces four distinct strategy variants through the combination of two key approaches for feature map mixing and two methods for label mixing calibration. The feature map mixing strategies include the Channel Max Position Detector and the Top K Channel Patches Detector, which are applied independently based on task requirements or data characteristics. For label mixing calibration, FMMix employs either the Feature Map Area Ratio or the Semantic Information Estimator to ensure accurate representation of each input sample contribution to the combined feature map.

The mixed-sample data augmentation process combines two input samples \((x_a, y_a)\) and \((x_b, y_b)\), where \(x_a, x_b,y_a, y_b\) are input images and it's corresponding labels, respectively. The augmented sample \((\tilde{x}, \tilde{y})\) is generated via a mixing function \(\tilde{g}\), the calculation of this process is outlined in \eqref{eq0}.
\begin{equation}
    (\tilde{x}, \tilde{y}) = \tilde{g}\big((x_a, y_a), (x_b, y_b)\big).
\label{eq0}
\end{equation}

In FMMix, feature maps are mixed follow by Eq \eqref{eq05}.
\begin{equation}
    \tilde{x}_l = (1 - M_C) \otimes h_l(x_a) + M_C \otimes h_l(x_b)
\label{eq05}
\end{equation}

where $M_C \in \mathbb{R}^{c \times h \times w}$ represents a channel-specific binary mask selected from a square patch centered at the maximum value position, and $\otimes$ denoted for the Hadamard product operation. The mixed label \(\tilde{y}\) is computed as a weighted combination of the original labels calculated by $\tilde{y} = p_a y_a + p_b y_b$ with \(p_a\) and \(p_b\) are mixing coefficients. FMMix introduces two key strategies for selecting informative regions from feature maps, which is the Channel Max Position Detector and the Top K Channel Patches Detector. For the Channel Max Position Detector, the position of maximum value in each channel is identified using. Calculated as in Eq \eqref{eq1} with \(h_l(x_b)\) at layer $l$, and $I_m$ denoted for the maximum value of each vector map.
\begin{equation}
    I_m = argmax_c(h_l(x_b))
\label{eq1}
\end{equation}

A random factor $\lambda \sim \text{Uniform}(0, 0.5)$, sampled from a uniform distribution between 0 and 0.5, determines the relative area of the patch to be extracted, a square patch centered at the maximum value position is then selected, to calculate the $M_c$ values. For label mixing calibration, FMMix proposes two alternative approaches to calculate the contribution ratios $p_a$ and $p_b$ with Feature Map Area Ratio ($p_{bm}$) and the Semantic Information Estimator ($p_{bs}$). In the Feature Map Area Ratio method, the contribution ratio of sample $b(p_{b})$ is computed as in Eq \eqref{eq2}.
\begin{equation}
    p_{bm} = \frac{1}{chw} \sum_{i=0}^{chw} M_{C_i}
\label{eq2}
\end{equation}
where $M_{C_i}$ denoted as an element in $i$-th of the $M_c$. This alternative approach calculates the contribution ratios based on semantic information in each channel, the feature map is first normalized, then semantic information percentages are computed and the contribution ratio of sample $b$ got calculated, these operation are expressed as follows with $CW$ symbolize for channel wise.
\begin{equation}
    p_{bs} = \frac{1}{C} \sum_{i=0}^{chw} \text{per\_semantic}_i \times M_{C_i}
\end{equation}
where
\begin{equation}
    \text{per\_semantic} = \frac{\text{shift}_{x_b}}{\sum_{\text{CW}}   \left(\text{shift}_{x_b}\right)}  
\end{equation}

\begin{equation}
    \text{shift}_{x_b} = h_1(x_b) - \min_{\text{CW}}(h_1(x_b))
\end{equation}

With $min_{CW}(\cdot)$ and $sum_{CW}(\cdot)$ used to find the min value and calculate sum across channel, respectively. For both methods, once $p_b$ is determined, the contribution ratio of sample a is computed as in Eq \eqref{eq3}.
\begin{equation}
    p_a = 1 - p_b
\label{eq3}
\end{equation}

These formulations enable FMMix to generate mixed samples that are semantically richer and more informative, facilitating the model ability to learn complex patterns and spatial relationships more effectively. In our work, we employed FMMix to enhance the classification performance on eye-tracking data for more effective approach for leveraging both spatial and semantic information, leading to superior performance across multiple evaluation metrics.

\begin{figure*}[!ht]
    \centering
    \includegraphics[width=\linewidth]{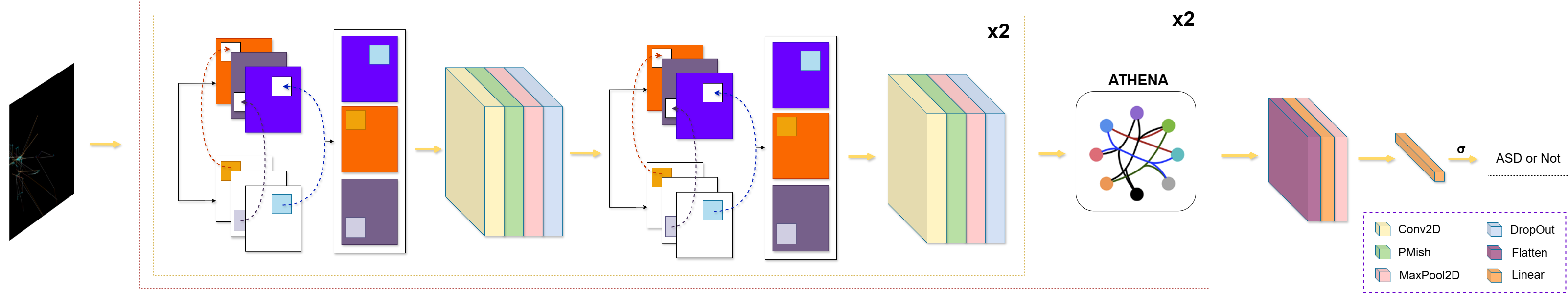}
    \caption{Proposal DNN Architecture with yellow denoted for the Conv2D layers, Green for Relu Activate Function, Pink represents for MaxPool2D layers, Blue is for Dropout layer, Purple for Flatten, and Orange is for the Linear layer.}
    \label{figm}
\end{figure*}

\subsection{Proposal DNN}
In this research, we propose a novel deep neural network (DNN) architecture for ASD classification by integrating feature-mixing augmentations, and ATHENA, designed to leverage convolutional feature extraction with hierarchical augmentation strategies, improving generalization and robustness. The proposed DNN model consists of multiple convolutional layers, activation functions, and pooling operations, followed by fully connected layers for classification. The convolutional feature extraction module comprises four convolutional layers, with the first two convolutional layers employ 16 filters with a kernel size of 3×3, followed by the Relu activation function, max-pooling, and dropout regularization, after that, these layers are complemented by ATHENA modules that facilitate adaptive feature learning. The subsequent two convolutional layers expand to 32 filters while maintaining Relu activation and dropout mechanisms to further enhance feature representation. The extracted feature maps are then processed through a fully connected layer comprising 256 neurons with Relu activation, followed by a final classification layer employing a sigmoid activation function to produce binary ASD classification. To enhance feature learning and mitigate overfitting, we utilized the FMMix augmentation to generates mixed samples by blending feature representations at intermediate layers, thereby increasing the diversity of training data. For further visualization Fig \eqref{figm} illustrated the model architecture.

\begin{figure*}[!ht]
    \centering
    \includegraphics[width=0.85\linewidth]{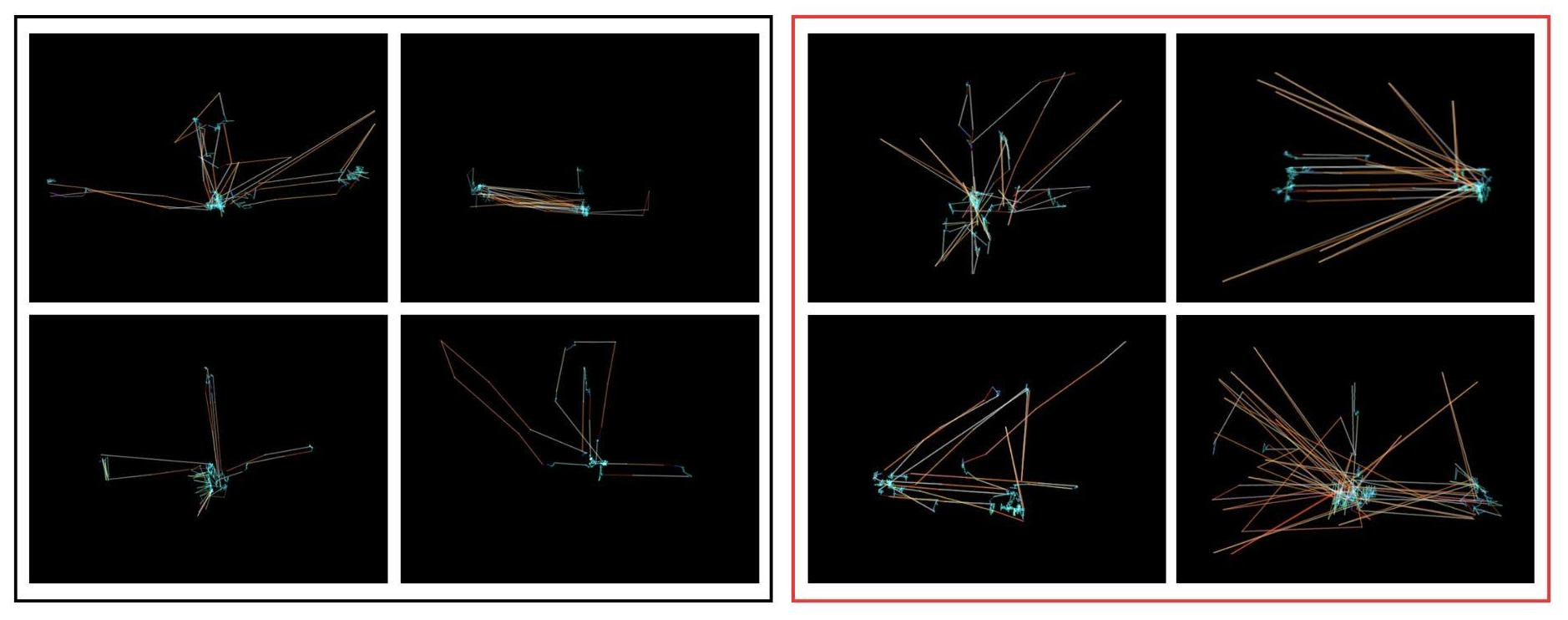}
    \caption{Samples data from Carette, R \textit{et al.} dataset with the red case on right representing the ASD class and black case at left showing the Non-ASD class.}
    \label{data1}
\end{figure*}
% To ensure efficient training, binary cross-entropy (BCE) loss is used as the primary loss function, with additional computation loss components derived from FMMix augmentation to regulate feature consistency. The Adam optimizer with a learning rate of \(10^{-3}\) is employed, incorporating weight decay for regularization. Data augmentation techniques, including rotation, shear, and zoom transformations, are applied to further enhance generalization and prevent overfitting.

\section{Material and Processing}
\subsection{Dataset}
\subsubsection{Carette, R \textit{et al.} Dataset}
For this research, we utilized a control dataset provided by Carette, R \textit{et al.} \cite{b19} to compare the patterns of eye movement among neurotypical individuals and those with autism performing similar visual tasks by using this dataset. It contains 547 images of eye-tracking paths (Eye Tracking SP) taken from eye-tracking (ET) data collected from 59 subjects. The experiment used visual stimuli specifically designed for people with autism. The device used was the SMI RED eye tracker. The author visualized eye tracking paths as three components as velocity, acceleration, and jerk. Red, Green, and Blue channels are assigned to velocity, acceleration, and jerk respectively. Higher values in each component result in a higher intensity of the associated color. By visualizing saccade paths as images, we aim to classify differences in eye movement patterns that are difficult to discern from raw eye-tracking data. These visual representations allow our machine learning models to learn more robust and informative features for analyzing gaze patterns, Fig \eqref{data1} illustrated sample images from the dataset between two classes.

\subsubsection{ASD-EyeMov Dataset} \label{sec2}
According to Duani \textit{et al.} (MMSys'19) \cite{b20}, eye movement recordings were obtained from 14 children with autism spectrum disorders (ASD) and 14 children with typically developing (TD). In this dataset, 300 images are presented across seven thematic categories to facilitate the analysis of disparities between the two groups when they are exposed to diverse visual stimuli.

Using a Tobii T120 eye tracker to record the (x, y) coordinates of the participants' gaze, both fixations and saccades between fixations were recorded. For each image presentation, data were sampled at 120 Hz for a period of three seconds. Consequently, raw data streams are made up of a series of coordinates (x, y) and associated timestamps.

\begin{figure*}[!ht]
    \centering
    \includegraphics[width=0.85\linewidth]{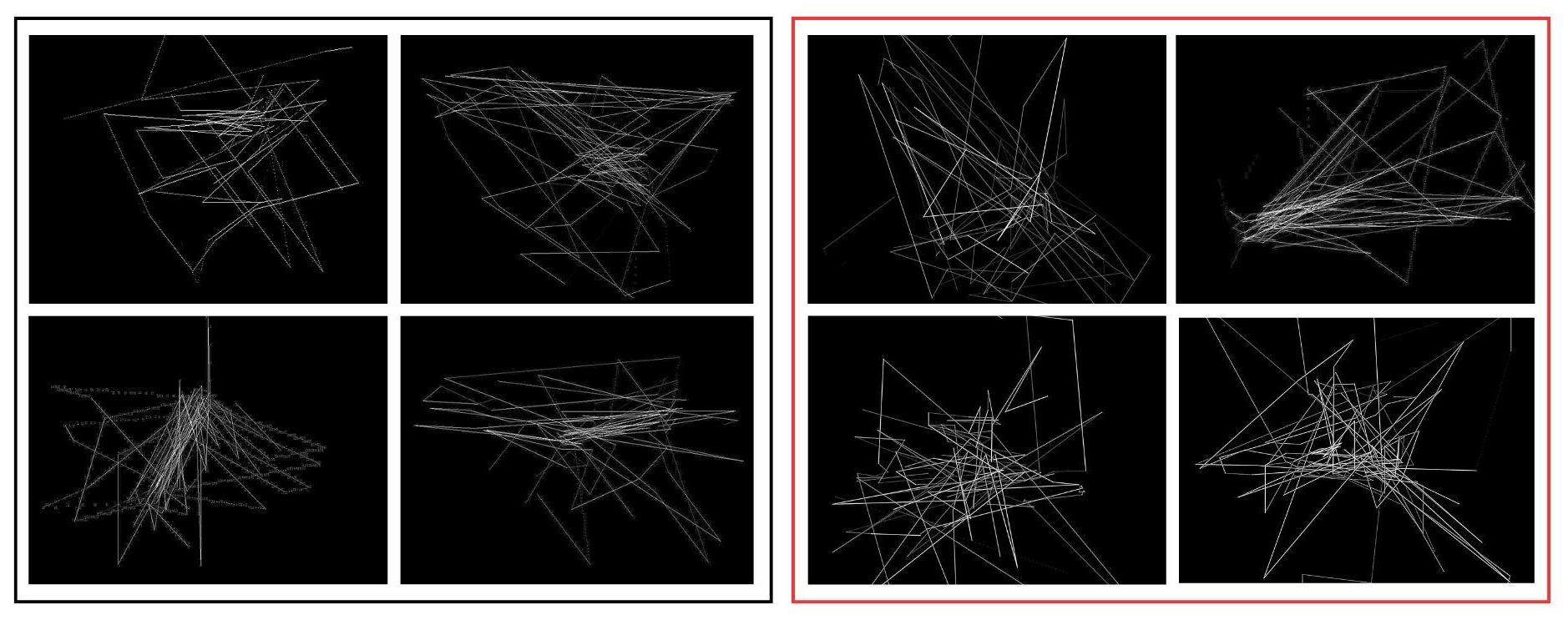}
    \caption{Oculomotor trajectory mapping creation from the ASD-Eyemov dataset with black case on left is for ASD and the red case of right is for the Non-ASD class}
    \label{figmov}
\end{figure*}

\subsection{Data preprocessing}
\subsubsection{Oculomotor Trajectory Mapping}
In this research, we utilized an oculomotor kinematics method that transcends conventional representation methods by incorporating multidimensional temporal-spatial data encoding since the dataset in Sec \eqref{sec2} is in static format. By that, in this research we proposed a method to generate topological manifolds from the dataset that capture the complex dynamical properties of saccadic and fixation behaviors by employing a sequential transformation operator where raw spatio-temporal coordinates undergo dimensional parsing through a string decomposition function \(\mathcal{S}\) that transforms \(\mathbb{R}\) into \(\mathbb{R}^4\). For each gaze fixation vector \(\vec{\delta}_i\) within the trajectory sequence, the data are structured as a quadruplet as presented in \eqref{eqd}.
\begin{equation}
    \vec{\delta}_i = (idx_i, x_i, y_i, d_i)
\label{eqd}
\end{equation}
Where $idx_i \in \mathbb{Z}^+$ represents the temporal sequence identifier, $x_i, y_i$ denotes the Cartesian projection of the three-dimensional gaze vector onto the visualization plane, and $d_i \in \mathbb{R}^+$ quantifies the fixation duration measured in milliseconds. To enhance perceptual discriminability in the visualization space, we applied an affine transformation to the temporal component using a min-max normalization operator \(\mathcal{N}\) that maps \(\mathbb{R}^+\) to the unit interval, the operation is outlined in \eqref{eqaa}.
\begin{equation}
    \mathcal{N}(d_i) = \frac{d_i - \min\limits_{j \in \Omega} d_j}{\max\limits_{j \in \Omega} d_j - \min\limits_{j \in \Omega} d_j}
\label{eqaa}
\end{equation}
where $\Omega = \{1,2,\ldots,n\}$ represents the index set of all fixation points. Afterward, a piece-wise linear interpolation function was utilized to transform pairs of points from \(\mathbb{R}^2\) into continuous functions in \(\mathcal{C}^0(\mathbb{R}^2)\), creating a continuous trajectory representation. The opacity function \(\alpha: \mathcal{D} \times \mathcal{D} \mapsto [0,1]\) is defined as in Eq \eqref{eqa}.
\begin{equation}
    \alpha(\vec{\delta}_i, \vec{\delta}_{i+1}) = d'_i
\label{eqa}
\end{equation}

This establishes a bijective mapping between the normalized temporal domain and the visual transparency domain, effectively encoding the multidimensional gaze behavior into a visually interpretable representation. After that, the visualization operation is rendered in a non-Euclidean perceptual space with coordinate normalization to the obtained dataset. By this, we proposed a visual representation of oculomotor behavior patterns that preserves both the geometric properties of the spatial trajectory and the statistical moments of the temporal distribution. In our research, we utilized this method to the ASD-Eyemov for the oculomotor trajectory mapping creation, the results is shown in Fig \eqref{figmov}.

\subsubsection{Data Normalization and Augmentation}
The first step in the process is pre-processing the input images, which includes two operations that convert RGB to grayscale and resizing the images to fit the machine learning model. The input images have three colors, as mentioned above, with a size of 640x480 pixels (Red, Green, and Blue). We use NSTC Eq \eqref{gs} to convert the RGB image to grayscale, where R, G, and B represent the pixel data for the Red, Green, and Blue colors, respectively. Afterwards, the image size must be reduced from $640 \times 480$ to $100 \times 100$ to reduce the computational burden, thus resulting in a faster computation.
\begin{equation}
     Grayscale = \frac{R}{0.299} + \frac{G}{0.587} + \frac{B}{0.114} 
\label{gs}
\end{equation}

To address the issue of insufficient original data and the risk of overfitting, we augmented the dataset by applying several transformations to the images with rotation of 10\%, 20\% in shear, and zoom of 20\%.

\section{Experiment Environments}
\subsection{Environments}
The experiments were conducted in a computational environment running Windows 11 Pro with an Intel Core i5-13400F processor and 32GB RAM. For GPU acceleration, we utilized an NVIDIA GeForce RTX 4070Ti SUPER graphics card. The software stack included Python 3.12.4, PyTorch 2.4.0, and CUDA 12.4.r12.4 toolkit for deep learning framework optimization. Furthermore, Table \eqref{tabexx} summarizes the experimental setup.

\begin{table}[!ht]
    \centering
    \caption{Experimental Setup.}
    \begin{tabular}{|c|c|}
    \hline
    \textbf{Category} & \textbf{Details} \\
    \hline
    \textbf{Hardware} &  \\
        CPU & Intel Core i5-13400F \\
        RAM & 32GB \\
        GPU & NVIDIA GeForce RTX 4070Ti SUPER \\
    \hline
    \textbf{Software} &  \\
        OS & Windows 11 Pro \\
        Language & Python 3.12.4 \\
        Pytorch version & PyTorch 2.4.0 \\
        Acceleration & CUDA 12.4.r12.4 \\
    \hline
    \end{tabular}
    \label{tabexx}
\end{table}

Model training was performed with the following hyperparameters to optimize convergence and generalization capability. With batch size of 12, learning rate at 0.001, AdamW selected as the optimizer and trained for 100 epochs. The Binary Cross-Entropy loss function (BCE) was utilized as the optimization criterion, appropriate for the binary classification nature of ASD detection.

\subsection{Evaluation Metrics}

\begin{table*}[h]
\centering
\caption{Evaluation results from the proposal DNN tested on two datasets.}
\begin{tabular}{|c|c|c|c|c|c|c|}
\hline
Model & AUC Score & Accuracy & PPV & NPV & Sensitivity & Specificity \\ \hline
Mahmoud E.'s Dataset & 99.80\% & 98.45\% & 98.14\% & 96.86\% & 98.01\% & 98.74\% \\ \hline
ASD-EyeMov Dataset & 99.86\% & 98.56\% & 99.04\% & 98.07\% & 98.10\% & 99.02\% \\ \hline
\end{tabular}
\label{tab123}
\end{table*}

\begin{table*}[!ht]
\centering
\caption{Comparison among different works using Dataset 1.}
\begin{tabular}{|c|c|c|c|c|c|c|}
\hline
Model & Best AUC & Accuracy & PPV & NPV & Sensitivity & Specificity \\ \hline
\textbf {Proposal model} & \textbf {99.80\%} & \textbf {98.45\%} & \textbf {98.14} & \textbf {96.86\%} & \textbf {98.01\%} & \textbf {98.74\%} \\ \hline
Mumenin, N. \textit{et al.} \cite{d21} & - & 98.12\% & 97.89\% & - & 99.03\% & - \\ \hline
Akter, T. et al. \textit{et al.} \cite{d31} & 71.5\% & 74.2\% & - & - & 74.2\% & 68.8\% \\ \hline
Kanhirakadavath, M. \textit{et al.} \cite{d32}& 97.0\% & - & 94.46\% & 90.06\% & 93.28\% & 91.38\% \\ \hline
Ahmed, Z. \textit{et al.} \cite{d33} & 98.0\% & 98.33\% & 98.11\% & 97.25\% & 97.25\% & 98.94\% \\ \hline
Zhao, Z. \textit{et al.} \cite{d34} & 86.0\% & 84.62\% & - & - & 89.47\% & 80.0\% \\ \hline
\end{tabular}
\label{tabd1}
\end{table*}

\begin{table*}[!ht]
\centering
\caption{Comparison among different works using Dataset 2.}
\begin{tabular}{|c|c|c|c|c|c|c|}
\hline
Model & AUC Score & Accuracy & PPV & NPV & Sensitivity & Specificity \\ \hline
\textbf{Proposed model}  & \textbf{99.86\%} & \textbf{98.56\%} & \textbf{99.04\%} & \textbf{98.07\%} & \textbf{98.10\%} & \textbf{99.02\%} \\ \hline
Mumenin, N. \textit{et al.} \cite{d21} & - & 96.83\% & 95.43\% & 96.13\% & 99.87\% & 99.84\% \\ \hline
Colonnesem F. \textit{et al.} \cite{d22} & - & 94.35\% & 93.2\% & - & 93.1\% & - \\ \hline
Zhou, W. \textit{et al.} \cite{d23} & - & 96.78\% & 96.89\% & - & 96.29\% & - \\ \hline
Liaqat, S. \textit{et al.} \cite{d24} & 64.0\% & 60.27\% & 67.81\% & - & 75.0\% & 40.0\% \\ \hline
Tao, Y. \textit{et al.} \cite{d25} & 57.9\% & 57.9\% & 56.3\% & - & 59.2\% & 56.6\% \\ \hline
\end{tabular}
\label{tabd2}
\end{table*}

For this research, we utilized multiple evaluation metrics for comprehensively assess model performance from various statistical perspectives. Firstly is Area Under the Receiver Operating Characteristic Curve (AUC), this metric provided a threshold-independent performance measure, representing the probability that the classifier will rank a randomly chosen positive instance higher than a randomly chosen negative instance. Also with Accuracy offers an intuitive overall performance measure across all predictions, while Positive Predictive Value (PPV) and Negative Predictive Value (NPV) represents the proportion of positive predictions that are correct, indicating the reliability of a positive ASD diagnosis and indicates the proportion of negative predictions that are correct, reflecting the reliability of ruling out ASD, respectively. Sensitivity is particularly important in ASD screening to ensure minimal missed cases requiring intervention, while Specificity prevents unnecessary referrals and psychological burden from false positive results. Together, these metrics establish a rigorous evaluation framework aligned with clinical priorities in ASD diagnosis. The calculation of the evaluation metric are outlined as in Eq \eqref{eqm1} - Eq \eqref{eqm2}.
\begin{equation}
    \text{AUC} = \int_{0}^{1} \text{PPV}(t) \times \text{NPV}'(t) dt
\label{eqm1}
\end{equation}

\begin{equation}
    \text{Accuracy} = \frac{\text{TP} + \text{TN}}{\text{TN + TP + FP + FN}} \times 100 
\end{equation}

\begin{equation}
    \text{PPV} = \frac{\text{TP}}{\text{TP} + \text{FP}} \times 100 
\end{equation}

\begin{equation}
    \text{NPV} = \frac{\text{TN}}{\text{TN} + \text{FN}} \times 100
\end{equation}

\begin{equation}
    \text{Sensitivity} = \frac{\text{TP}}{\text{TP} + \text{FN}} \times 100
\end{equation}

\begin{equation}
    \text{Specificity} = \frac{\text{TN}}{\text{TN} + \text{FP}} \times 100
\label{eqm2}
\end{equation}
where \textit{t} respresent for threshold, TP, FP, TN, FN, denoted for True prediction, False prediction, True negative, and False negative, respectively.

\section{Experiment Results}
The proposed DNN model shown an robust performance across both datasets utilized in this study. Table \ref{tab123} presents a comprehensive overview of the evaluation metrics obtained through rigorous testing on the Carette, R \textit{et al.} dataset and the ASD-EyeMov dataset.

On the Carette, R \textit{et al.}, our model achieved an AUC score of 99.80\%, indicating near-perfect discrimination between ASD and neurotypical classes. The test accuracy was similarly impressive at 98.45\%, with a PPV of 98.14\% and NPV of 96.86\%. The model demonstrated robust sensitivity and specificity values of 98.01\% and 98.74\% respectively, confirming its ability to correctly identify both ASD and non-ASD cases with high reliability.

Performance on the ASD-EyeMov dataset was more robust, with the model attaining an AUC score of 99.86\%, reflecting exceptional discriminative capability. The test accuracy reached 98.56\%, while PPV and NPV values were 99.04\% and 98.07\% respectively. Sensitivity measurements indicated that 98.10\% of actual ASD cases were correctly identified, and specificity results showed that 99.02\% of non-ASD cases were accurately classified, these metrics shown that the model performance is more effective on the ASD-EyeMov dataset, particularly in terms of positive predictive value and specificity. The consistent performance across two independent datasets provides compelling evidence for the robustness and generalizability of our approach. The high sensitivity values across both datasets of 98.01\% and 98.10\% demonstrate the model effectiveness in identifying individuals with ASD, which is particularly critical in clinical applications where early detection facilitates timely intervention. Furthermore, the high specificity measurements at 98.74\% and 99.02\% indicate a low false positive rate, thus minimizing the risk of misdiagnosis and unnecessary interventions.

These results shown the combination adaptive threshold hypergraph enhanced neural architecture, and feature map mixing has proven highly effective for extracting discriminative features from eye-tracking data. The exceptional performance metrics suggest that the proposed methodology successfully captures the subtle yet distinctive patterns in visual attention and gaze behavior that differentiate individuals with ASD from neurotypical controls. The marginal performance variations between datasets may be attributed to differences in data collection protocols, participant demographics, or the specific visual stimuli employed during eye-tracking assessments.

\section{Discussion}
\subsection{Comparisons with other works}
The experimental results demonstrate the superior performance of our proposed DNN architecture in multiple metrics in both data sets, the results are summarized in Table \eqref{tabd1}, and Table \eqref{tabd2} across both dataset. when comparing our model with existing approaches, we observe significant improvements in classification performance.

\begin{figure*}[h]
    \centering
    
    \begin{subfigure}{\linewidth}
        \centering
        \includegraphics[width=0.9\linewidth]{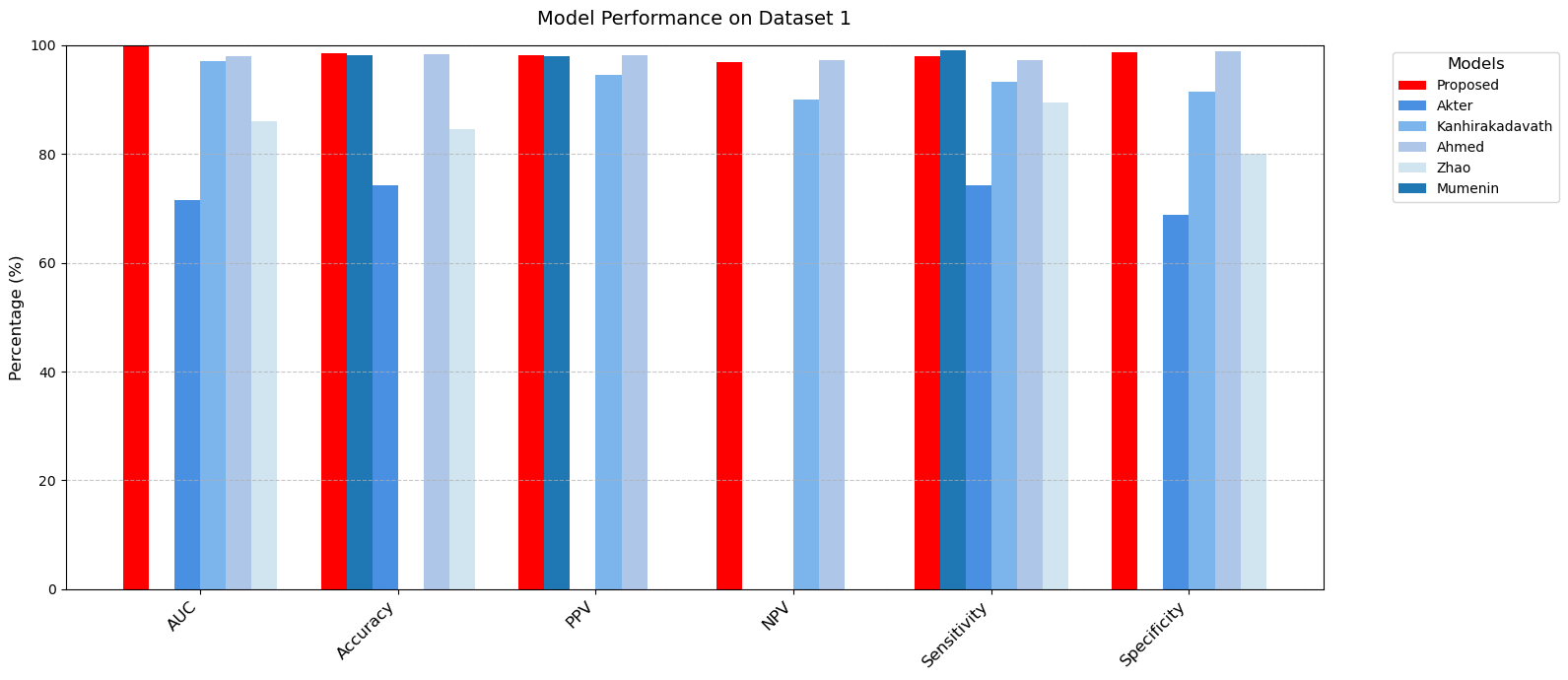}
        \caption{Results comparisons from Dataset 1}
        \label{figdd1}
    \end{subfigure}
    
    \vspace{0.5cm}
    
    \begin{subfigure}{\linewidth}
        \centering
        \includegraphics[width=0.9\linewidth]{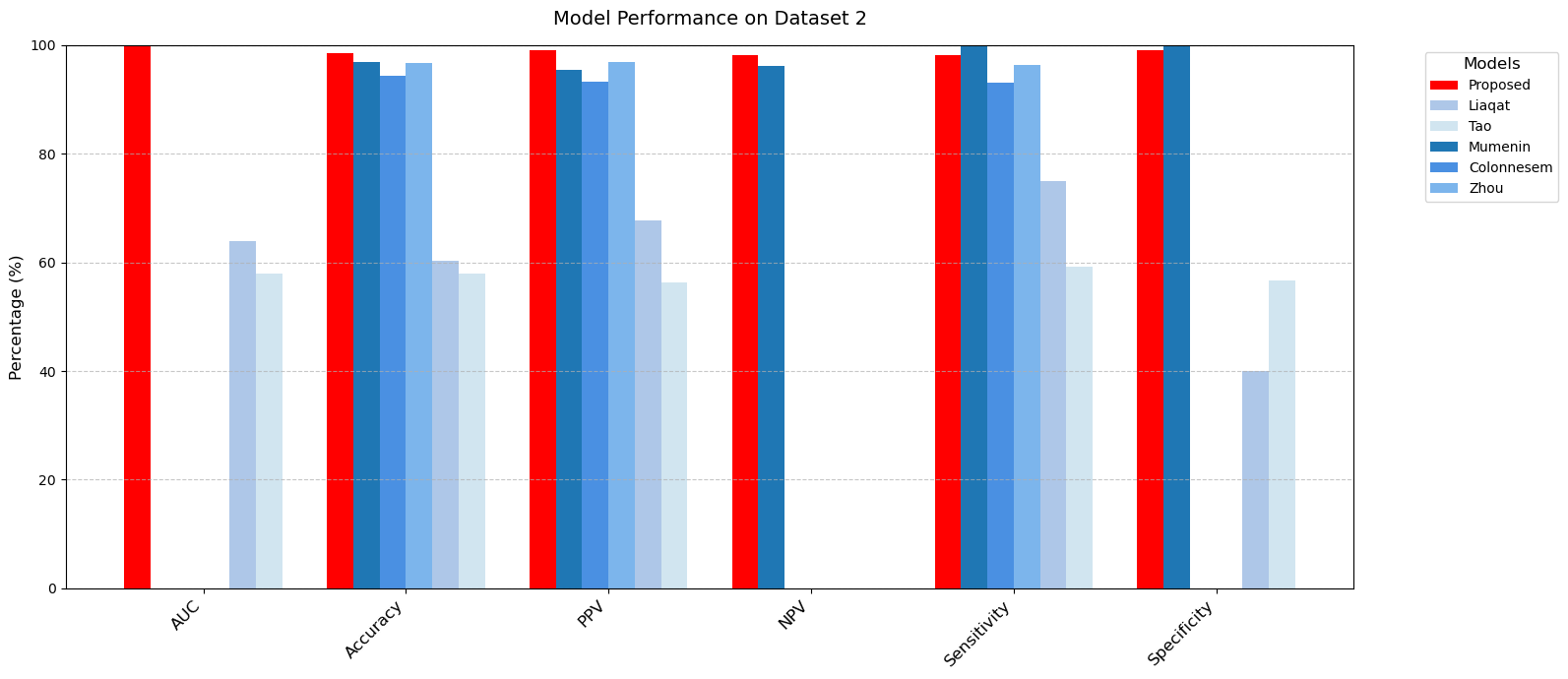}
        \caption{Results comparisons from Dataset 2}
        \label{figdd2}
    \end{subfigure}
    
    \caption{Performance comparison results across two datasets.}
    \label{fig:comparison}
\end{figure*}

Table \eqref{tabd1} compares our model against five other approaches using the ASD-EyeMov dataset, which contains eye movement recordings from 14 children with ASD and 14 typically developing children. Our model achieves an outstanding AUC score of 99.86\%, significantly outperforming the next best reported scores from Liaqat \textit{et al.} at 64.0\% and Tao \textit{et al.} at 57.9\%. This improvement got over 35 percentage points suggests our approach captures patterns in the data that other methods miss entirely. However, it's worth noting that three competing approaches do not report AUC scores, making the comparison incomplete. For accuracy, our model achieved 98.56\% outperforms all competitors, with the closest being Mumenin \textit{et al.} at 96.83\% and Zhou \textit{et al.} at 96.78\%. This represents an improvement of approximately 1.73-1.78 percentage points over the next best methods, which is substantial in a field where improvements often come in fractions of a percent. Our model's PPV got 99.04\% exceeds all other approaches, with Zhou \textit{et al.} achieved 96.89\% being the closest competitor. This exceptionally high PPV indicates superior precision in ASD detection, with very few false positives, the NPV of 98.07\% also outperforms the only other reported value from Mumenin \textit{et al.} at 96.13\%, suggesting better reliability in ruling out ASD. Interestingly, our sensitivity at 98.10\% is slightly lower than Mumenin \textit{et al.} 99.87\%, indicating their model might detect slightly more true ASD cases. However, this comes at the cost of lower PPV, suggesting potential over-diagnosis. The specificity of our model return with 99.02\% is excellent but slightly lower than Mumenin \textit{et al.} at 99.84\%, though the difference is minimal. The performance differences are particularly striking when comparing against less effective methods like Liaqat \textit{et al.} and Tao \textit{et al.}, where our model shows improvements of over 35\% points across multiple metrics.

\begin{table*}[!ht]
\centering
\caption{Comparison among different activation functions using Dataset 1.}
\begin{tabular}{|c|c|c|c|c|c|c|}
\hline
Model & AUC Score & Accuracy & PPV & NPV & Sensitivity & Specificity \\ \hline
PMish & 96.97\% & 92.66\% & 93.98\% & 91.90\% & 86.92\% & 96.38\% \\ \hline
Relu & 96.75\% & 92.18\% & 90.66\% & 93.15\% & 89.35\% & 94.03\% \\ \hline
Silu & 97.31\% & 93.52\%  & 94.57\% & 92.90\%  & 88.62\% & 96.70\%  \\ \hline
LRelu & 97.84\% & 93.80\%  & 95.79\% & 92.68\%  & 88.14\% & 97.48\% \\ \hline
Mish & 96.84\% & 92.09\%  & 91.67\% & 92.34\%  & 87.89\% & 94.81\% \\ \hline
Swish & 97.04\% & 92.66\%  & 93.75\% & 92.03\%  & 87.17\% & 96.23\% \\ \hline
\end{tabular}
\label{act1}
\end{table*}

\begin{table*}[!ht]
\centering
\caption{Comparison among different activation functions using Dataset 2.}
\begin{tabular}{|c|c|c|c|c|c|c|}
\hline
Model & AUC Score & Accuracy & PPV & NPV & Sensitivity & Specificity \\ \hline
PMish & 99.58\% & 97.12\% & 98.54\% & 95.73\% & 95.73\% & 98.54\% \\ \hline
Relu & 99.83\% & 98.08\% & 97.65\% & 98.52\% & 98.58\% & 97.56\% \\ \hline
Silu & 99.25\% & 97.00\%  & 97.83\% & 96.16\%  & 96.21\% & 97.80\%  \\ \hline
LRelu & 99.56\% & 98.20\%  & 98.34\% & 98.05\%  & 98.10\% & 98.29\% \\ \hline
Mish & 99.60\% & 97.00\%  & 98.30\% & 95.72\%  & 95.73\% & 98.29\% \\ \hline
Swish & 99.32\% & 97.36\%  & 97.85\% & 96.86\%  & 96.92\% & 97.80\% \\ \hline
\end{tabular}
\label{act2}
\end{table*}

Table \eqref{tabd2} presents a comprehensive comparison of our proposed model against five existing approaches on the Carette \textit{et al.} dataset, which contains eye-tracking paths from 59 subjects. Our proposed model achieves an AUC score of 99.80\%, achived highest results compared to all methods, this near-perfect AUC indicates exceptional discriminative ability between ASD and non-ASD classes. The closest competitor in terms of AUC is Ahmed \textit{et al.} with 98.0\%, representing a small but meaningful improvement of 1.8\% points. The substantial gap between our model and lower-performing approaches like Akter \textit{et al.} with 71.5\% and Zhao \textit{et al.} with 86.0\% highlights the effectiveness of our novel architecture. In terms of accuracy with 98.45, our model slightly outperforms Ahmed \textit{et al.} at 98.33 and Mumenin \textit{et al.} at 98.12. Our model demonstrates more balanced performance across all metrics, suggesting robust generalization capabilities rather than optimization for a single metric. The PPV of 98.14\% is particularly noteworthy, as it nearly matches Ahmed \textit{et al.} at 98.11\% while surpassing Kanhirakadavath \textit{et al.} with 94.46\%. This high PPV indicates that when our model predicts ASD, it has a very high probability of being correct, which is crucial for clinical applications to minimize false alarms. For NPV, our model achived 96.86 outperforms both Kanhirakadavath \textit{et al.} with 90.06\% and Ahmed \textit{et al.} with 97.25, suggesting superior reliability when ruling out ASD. Regarding sensitivity, our model got 98.01 performs strongly but is slightly below Mumenin \textit{et al.} at 99.03. This small difference suggests our model might miss slightly more ASD cases, though the difference is minimal. Our specificity take in with 98.74 outperforms most competitors but falls marginally short of Ahmed \textit{et al.} with 98.94, indicating nearly equivalent performance in correctly identifying non-ASD cases.

This suggests that our integrated approach of ATHENA, and FMMix fundamentally changes the model's ability to identify relevant patterns in the eye-tracking data. Both tables demonstrate that our proposed architecture not only achieves state-of-the-art performance but also maintains remarkable balance across all evaluation metrics, suggesting a robust and clinically relevant approach to ASD classification using eye-tracking data, for better visualization Fig \eqref{figdd1} and Fig \eqref{figdd2} outline the results from the Dataset 1, and the Dataset 2.

\subsection{Activate Function Comparisons}
In this study, we conducted a comparative analysis of six activation functions as PMish, ReLU, SiLU, Leaky ReLU (LReLU), Mish, and Swish within the CNN architecture, assessing their performance across two distinct datasets. The results, detailed in Table \eqref{act1} and Table \eqref{act2} for Dataset 1 and Dataset 2, respectively.

For Dataset 1, the LReLU activation function consistently outperformed all other functions across every metric. It achieved an AUC score of 97.84\%, an Accuracy of 93.80\%, a PPV of 95.79\%, an NPV of 92.68\%, a Sensitivity of 88.14\%, and a Specificity of 97.48\%. In contrast, PMish recorded slightly lower values, with an AUC of 96.97\% and an Accuracy of 92.66\%, while other functions such as SiLU achieved AUC of 97.31\% and Swish with AUC 97.04\% remained competitive but fell short of LReLU performance. These findings indicate that LReLU offers a superior balance of sensitivity and specificity for Dataset 1, making it the most effective activation function in this context.

The performance for Dataset 2  was more varied. Here, ReLU achieved the highest AUC score of 99.83\%, demonstrating exceptional discriminative power, while LReLU recorded the highest Accuracy at 98.20\%. PMish stood out with a PPV and Specificity both at 98.54\%, though its Sensitivity of 95.73\% was notably lower than that of ReLU 98.58\% and LReLU with 98.10\%. Mish also exhibited strong performance, with an AUC of 99.60\% and PPV and Specificity values comparable to PMish. This variability suggests that the optimal activation function for Dataset 2 depends on the specific classification priorities. For applications prioritizing the minimization of false positives, PMish high Specificity is advantageous, whereas ReLU or LReLU may be preferred when maximizing true positive detection is critical.

% Overall, the analysis highlights distinct performance profiles across the two datasets. LReLU dominance on Dataset 1 contrasts with the metric-specific strengths of ReLU, LReLU, and PMish on Dataset 2. These results emphasize the importance of tailoring the choice of activation function to the dataset characteristics and the intended application, ensuring alignment with the desired balance of performance metrics.

\subsection{Experiments Comparisons with Athena Module}
In this subsection, a comparative analysis of the CNN architecture with and without the integration of the ATHENA module, performance evaluation is outlined in Table \eqref{at1}. The baseline CNN model, without the ATHENA module, achieved an AUC score of 96.75\%, an Accuracy of 92.18\%, a PPV of 90.66\%, an NPV of 93.15\%, a Sensitivity of 89.35\%, and a Specificity of 94.03\%. Upon incorporating the ATHENA module, these metrics exhibited significant improvements, with the AUC score increasing to 98.39\%, Accuracy to 94.76\%, PPV to 93.66\%, NPV to 95.45\%, Sensitivity to 92.98\%, and Specificity to 95.31\%.

\begin{table*}[!ht]
\centering
\caption{Ablation study on Athena Layer.} 
\begin{tabular}{|c|c|c|c|c|c|c|}
\hline
Model & AUC Score & Accuracy & PPV & NPV & Sensitivity & Specificity \\ \hline
CNN Model & 96.75\% & 92.18\% & 90.66\% & 93.15\% & 89.35\% & 94.03\% \\ \hline
CNN Model with Athena & 98.39\% & 94.76\% & 93.66\% & 95.45\% & 92.98\% & 95.31\% \\ \hline
\end{tabular}
\label{at1}
\end{table*}

The ATHENA module enhances the CNN model by employing an adaptive hypergraph framework to dynamically capture high-order feature correlations within the eye-tracking. The notable rise in AUC from 96.75\% to 98.39\% indicates a substantial enhancement in the model robustness. Additionally, the improvements in Accuracy, PPV, and NPV reflect a higher precision and reliability in predictions, effectively reducing both false positives and false negatives. The gains in Sensitivity and Specificity are particularly impactful, as they demonstrate that the model with ATHENA excels at identifying toddlers with ASD while accurately excluding the disorder in neurotypical cases. This balance is crucial for clinical applications, where early detection can enable timely interventions, and minimizing misdiagnoses helps avoid unnecessary stress and resource expenditure. These findings highlight the effectiveness of the ATHENA module in augmenting the baseline CNN architecture, allowing it to better capture subtle, clinically significant gaze patterns indicative of ASD. The consistent enhancements across all evaluated metrics underscore the robustness and generalizability conferred by adaptive hypergraph-based feature modeling. Consequently, this advancement offers considerable potential for improving early ASD screening tools, paving the way for more reliable and non-invasive diagnostic solutions in clinical settings.

\section{Conclusion and Future work}
In conclusion, this research introduces a novel DNN architecture, enriched by the ATHENA hypergraph structure, and FMMix data augmentation, representing a significant stride forward in automated ASD detection. Achieving a superior AUC of 99.86\% on the ASD-EyeMov dataset, a test accuracy of 98.56\%, and well-balanced PPV, NPV, sensitivity, and specificity values exceeding 98\% on both Dataset 1 and Dataset 2, compared to existing methodologies, underscores the efficacy and robustness of our strategy. The integration of these innovative components enables our model to discern subtle yet vital patterns within eye-tracking data, resulting in more reliable and precise ASD classification. Moving forward, it is crucial to conduct extensive validation on larger, more diverse, and independent datasets to ascertain generalizability within clinical settings and datasets with varied demographics. Further enhancing the integration of multi-modal data, such as behavioral insights or neuroimaging data, may enhance diagnostic precision. Moreover, investigating the potential for personalized ASD screening, tailored to individual gaze patterns, could enable targeted screening for improved patient care. Finally, the development of a real-time model implementation may facilitate immediate assessment of ASD risk, allowing for early intervention and better patient outcomes. Addressing these facets of future research may further refine our approach, bolstering the development of effective tools for timely detection and intervention.

\section*{Declaration of competing interest}
The authors declare that they have no known competing financial interests or personal relationships that could have appeared to influence the work reported in this paper.

\section*{Acknowledgements}
This research is supported by International School, Vietnam National University, Hanoi (VNU-IS).

\bibliographystyle{cas-model2-names}

% \section*{Data Availability}
% The dataset supporting this study is currently anonymized for the review process and will be released upon acceptance.

\appendix
\section{Theoretical Analysis of Adaptive Threshold Mechanism}

The theoretical foundation of our adaptive threshold mechanism is characterized by two key properties followed by gradient flow and adaptive receptive field formation. The gradient flow equation decomposes the learning process into two parallel pathways, with the first term, involving $\partial \theta_a / \partial \phi$, capturing how the threshold responds to changes in global feature context, while the second term, involving $\partial \theta_m / \partial \phi$, reflects sensitivity to dominant feature responses. This dual-path design ensures the threshold adapts to both overall feature distributions and salient local patterns, the overall representation outlined in Eq \eqref{tp} where $\tau = \sigma(W_2 \times \varphi(W_1 \times (\theta_a \oplus \theta_m)))$.

\begin{equation}
    \partial \lambda / \partial \phi = \epsilon \times (\partial \tau / \partial \theta_a \times \partial \theta_a / \partial \phi + \partial \tau / \partial \theta_m \times \partial \theta_m / \partial \phi).
\label{tp}
\end{equation}

where the partial derivatives decompose the gradient pathway into average and maximum pooling branches, this equation ensures smooth gradient propagation through both global context and salient feature pathways. For the adaptive receptive field, the dynamic threshold leads to an adaptive receptive field $R(v)$ for each vertex v, we define $R(v) = {u | d(f_v, f_u) < \lambda}$ where $d(\cdot)$ denotes the euclidean distance in feature space, $f_v, f_u \in \mathbb{R}^c$ denoted for the feature vectors in hypergraph structure of vertex $v$ and $u$, respectively. The adaptive receptive field defines a dynamic neighborhood system for each vertex based on feature similarity, this data-dependent connectivity allows the network to capture varying scales of feature dependencies and adapt to different input patterns where the threshold $\lambda$ is a learnable parameter that controls the extent of feature aggregation. The interplay between these properties allows the network to effectively balance local and global feature interactions through learnable thresholding, the threshold-gradient relationship operation can be defined by Eq \eqref{.fd}.

\begin{equation}
    \frac{\partial |R(v)|}{\partial \lambda} = \sum_{u \in V} \delta(\lambda - d(f_v, f_u))
\label{.fd}
\end{equation}

This relationship characterizes how changes in the threshold affect receptive field size. The summation over indicator functions provides a discrete measure of connectivity sensitivity, allowing the network to adjust its aggregation based on the learned threshold value.
The proposed method of adaptive thresholding with dynamic receptive fields creates a flexible framework for feature learning, the mechanism dynamic balances local and global feature interactions through learnable thresholds, while maintaining efficient computation through focused connectivity. This adaptive approach enables more nuanced feature aggregation compared to fixed-threshold methods, particularly in scenarios with varying scales of semantic information.

\printcredits

%% Loading bibliography style file

% Loading bibliography database
% \bibliography{cas-refs}

\end{document}